# Riconoscimento ortografico dell'apostrofo e delle espressioni polirematiche

*Algoritmi di correzione ortografica automatica basati su dizionari di grandi dimensioni*

Massimiliano Polito
Network Contacts - Via Olivetti 17 Molfetta (Ba)

## Abstract

**English**.
The work presents two algorithms of manipulation and comparison between strings whose purpose is the orthographic recognition of the apostrophe and of the compound expressions. The theory supporting general reasoning refers to the basic concept of EditDistance, the improvements that ensure the achievement of the objective are achieved with the aid of tools borrowed from the use of techniques for processing large amounts of data on distributed platforms.

**Italiano.**
Il lavoro presenta due algoritmi di manipolazione e confronto tra stringhe il cui scopo è il riconoscimento ortografico dell'apostrofo e delle espressioni polirematiche. La teoria a sostegno del ragionamento generale riferisce il concetto base di EditDistance, i perfezionamenti che garantiscono il raggiungimento dell'obbiettivo si realizzano con l'ausilio di strumenti mutuati dall'uso di tecniche di elaborazione di grandi moli di dati su piattaforma distribuita.

**INDICE DELLE TABELLE**

# 1. Introduzione

## 1.1 Scopo

Un sistema di controllo ortografico è un automa capace di confrontare rapidamente un testo con un *golden standard* composto da vocaboli scritti correttamente, la cui natura garantisce una base di confronto certa per *rivelare* la presenza di un errore di ortografia; il quale, una volta riconosciuto e localizzato nella frase, può lasciare il posto al vocabolo corretto.

Es:

**Frase**: *Lacqua score verso il basso*
**Correzione**: *L'acqua scorre verso il basso*

L'aspetto peculiare cui si riferisce il presente lavoro illustra la soluzione pensata per risolvere i principali errori di digitazione con particolare attenzione verso due aspetti particolari della lingua, per i quali non è sufficiente la sola presenza di un golden standard:

- sintagmi in cui è previsto l'apostrofo come in: *l'acqua, un'offerta*.
- espressioni polirematiche, ovvero sintagmi composti da termini ciascuno dei quali possiede un proprio significato specifico ma la cui coesistenza in sequenza esprime un concetto più ampio o molto diverso, come in: *Comunità Europea, Gruppo di Continuità, Conto Corrente*.

La costruzione del processo di controllo muove dal concetto di Edit Distance (ED) che è una delle possibili declinazioni della distanza di *Damerau-Levenshtein*[1], ma per intervenire con efficacia sui casi indicati sopra si rende necessaria la progettazione di adeguati algoritmi nei quali lo strumento di ED si adopera in accordo a modalità di trattamento del testo e a espansioni del dizionario, cosi da amplificare l'applicabilità del concetto di distanza, estendendola oltre il confine del singolo vocabolo.

## 1.2 Premessa

Per garantire la corretta applicazione del concetto di ED, si procede alla costruzione di un dizionario mutuando i vocaboli da un Corpus sulla cui idoneità non vi sia alcuna ambiguità. Il Corpus su cui cade la nostra scelta è **Paisà**, ampia collezione di testi autentici in lingua italiana tratti da Internet [1], la cui varietà assicura un dizionario sufficientemente ampio, con la possibilità di arricchimento con vocaboli specifici o gergali, relativi ad ambiti circoscritti: informatica, medicina, geologia e cosi via. Ciascun vocabolo è corredato con la propria frequenza assoluta nel Corpus, che verrà adoperata nel processo di riconoscimento ortografico. Il trattamento di un Corpus di grandi dimensioni è un procedimento la cui natura si adatta perfettamente alla capacità di calcolo tipica degli ambienti distribuiti e degli strumenti per l'elaborazione concorrente di una grande mole di dati [2], e proprio in virtù della loro duttilità è possibile

---

[1] La distanza di Levenshtein tra due stringhe A e B è il numero minimo di modifiche elementari che consentono di trasformare la A nella B

sfruttare la numerosità dei testi che fornisce materiale linguistico adeguato non solo a costruire un *golden standard* esauriente ma anche a evidenziare un'ulteriore caratteristica, fondamentale per gli algoritmi di correzione, ovvero la costruzione di liste [2] costituite da n-ple (con n>=2) di vocaboli che molto spesso si trovano scritti in sequenza; ad esempio:

*CASA, DI, PRODUZIONE,*

*CASA, DI, RIPOSO*

*PRESIDENZA, DEL, CONSIGLIO*

L'esistenza di queste n-ple ci consente di stabilire che il vocabolo *CASA* è seguito (nel Corpus) dai vocaboli: *DI*, *DEL*, *EDITRICE*, *SUA* ma anche *BIANCA*, *ROSADA* e cosi via con diverse frequenze di occorrenza. Per cui ogni vocabolo inteso come possibile testa sintagmatica, ha un insieme ammissibile di modificatori a cui può associarsi. Se è vero che esiste la coppia CASA -> DISCOGRAFICA, non esiste invece la coppia CASA -> LIMONE; il corpus fornisce un'evidenza empirica di ciò che è espresso nel *Vincolo della Verosimiglianza*, secondo il quale alcune parole hanno probabilità nulla di occorrere in coppia, altre hanno probabilità superiore a zero e tra queste alcune hanno probabilità superiori alla media [3].

### 1.2.1 LookUp - Vocabolo singolo

La correzione (*LookUp*) del singolo vocabolo prevede la costruzione di un dizionario che amplia il golden standard di riferimento, nella misura indicata dal valore di *maxEditDistance* che rappresenta il numero massimo di caratteri da sottrarre ad un dato vocabolo (elemento **generatore**) per ottenerne altri "vicini" ma ortograficamente non corretti [4]. Illustriamo il procedimento utilizzando un vocabolo di riferimento: "*acqua*" e due valori di *maxEditDistance:*

| maxEditDistance = 1 | acqua |
|---|---|
| | aqua |
| | acua |
| | cqua |
| | acqu |
| | acqa |

*Tabella 1.1 - Rimozione di un carattere*

Se il valore indicato per *maxEditDistance* è 2, si realizza la sequenza definita in Tabella 1.2, in cui si evidenzia come il procedimento, in aggiunta alle determinazioni individuate per *maxEditDistance=1* (Tabella 1.1), introduce nuove determinazioni, ottenute eliminando coppie di caratteri dal vocabolo di riferimento.

| maxEditDistance = 2 | acqua |
|---|---|
| | aqua |
| | acua |
| | cqua |
| | acqu |
| | acqa |
| | aqu |
| | cqu |
| | aua |
|   | cua |
|   | acq |
|   | aqa |
|   | cqa |
|   | acu |
|   | qua |
|   | aca |

*Tabella 1.2 - Rimozione di due caratteri*

Lo scopo delle rimozioni è la creazione di un vocabolario accessorio che chiameremo *Knowledge Base* (KB), di possibili errori da affiancare al termine corretto, l'intensità della rimozione ovvero il valore del parametro *maxEditDistance* è stabilito a priori, è caratteristico del sistema ed è determinante anche nella fase di correzione. La costruzione della KB è un procedimento costoso in termini di tempo e rientra nella fase preparatoria dell'ambiente, ovviamente maggiore è il valore di *maxEditDistance* e più onerosa diventa l'espansione ortografica con il conseguente aumento del tempo di preparazione del sistema.

Correggere un vocabolo significa individuare un elemento che sia *prossimo* al valore di input. La prossimità è calcolata in termini di distanza di *Damerau-Levenshtein* ed è un valore intero nell'intervallo [0, *maxEditDistance*]. Il sistema non può correggere ortografie che richiedono un numero di manipolazioni superiori al valore utilizzato per costruire la KB:

Sia *V* il vocabolo indicato in input, la correzione di *V* si ottiene dal confronto con la KB e distinguiamo due casi.

1. V è presente in KB come elemento generatore, significa che è ortograficamente corretto, la risposta del sistema è V.
2. V non è un elemento generatore, quindi non è ortograficamente corretto, si avvia un processo di confronto all'interno della KB per scegliere la stringa $S \in KB$ più vicina a V (distanza di *Damerau-Levenshtein*), il vocabolo che corregge V è l'elemento generatore di S.

Utilizziamo un vocabolo d'uso comune per illustrare il percorso della correzione: SMARTPHONE.

L'uso di parole in lingua inglese è accettato in quasi tutti i contesti, ma in particolare in quelli dove la colloquialità è caratterizzata da grande informalità: chat, post, tweet, messaggi su Whatsapp. Uno degli errori più comuni in questi casi, è la sostituzione di due o più caratteri con altri che ne rappresentano la pronuncia in italiano, per cui è ammissibile ipotizzare la traslazione SMARTPHONE -> SMARTFON, secondo *Damerau-Levenshtein* i due vocaboli hanno distanza pari a 3, perché per passare dal primo al secondo occorrono: due rimozioni (P, E) e una sostituzione (H con F). La procedura prende in input la parola SMARTFON e agisce in maniera simile al processo di costruzione della KB derivata dal dizionario, per cui da SMARTFON si eliminano progressivamente un numero di caratteri che va da 1 fino alla massima distanza di *edit* prevista per quel termine, stimata pari *al numero intero più prossimo a un terzo della lunghezza L del vocabolo*[2], nel caso in esempio si ha L=8 quindi *edit* = 3. Si ottengono determinazioni specifiche eliminando

---

[2] Nei procedimenti di correzione in genere la distanza di edit è fissa, pari al valore di *MaxEditDistance*, in base al quale è stata generata la KB. Questa pratica non tiene conto della lunghezza della stringa su cui il sistema agisce e rischia di ampliare in modo incongruo il bacino di possibili correzioni. Nella nostra implementazione scegliamo una distanza di *edit* variabile, calcolata sulla base della lunghezza della stringa; sia L la lunghezza della stringa allora

$$edit = Int(L * k) \qquad (1.1)$$

a rotazione da 1 a 3 caratteri dalla stringa in input e se una delle determinazioni ottenute è presente nella KB, l'elemento generatore ad essa associato (caratterizzato da una propria misura di distanza dal termine corretto) è il vocabolo che **probabilmente** corregge il termine in input. Naturalmente da generatori diversi potrebbero scaturire una o più determinazioni identiche; ad esempio *SMARTPHONE(1)* e *SMARTBOX(2)* avranno in comune *SMARTO*, termine indotto dalla cancellazione di caratteri tanto nella prima quanto nella seconda parola. Anche SMARTFON genera SMARTO, e quindi l'intersezione delle determinazioni individua in (1) e (2) almeno due elementi generatori; nel caso (1) la distanza è 3, nel caso (2) la distanza è 2. Ortograficamente SMARTFON è più vicino a *SMARTBOX*, semanticamente è esattamente il contrario ma il riconoscimento ortografico non tiene alcun conto della circostanza e quindi il generatore (2) è preferito al generatore (1); è nella filosofia del sistema prediligere la distanza minore e nei casi in cui c'è pareggio sulla misura della distanza, si adotta come discriminante la frequenza assoluta del generatore, calcolata nel Corpus da cui scaturisce il dizionario di base. Se l'ortografia in input fosse: *SMART FON* o *SMARTFONE* allora la correzione è (1) con una distanza pari a 3; (2) continua ad essere una correzione possibile ma con quella grafia la sua distanza è pari a 3: in pareggio con (1), in questo caso gioca un ruolo determinate la frequenza che per (1) è molto maggiore di (2), quindi la scelta del vocabolo che corregge l'ortografia di *V* ricade su (1). Naturalmente la lookUp singola non tiene alcun conto del contesto frastico, si limita a verificare un vocabolo senza alcuna contestualizzazione; ma è il procedimento in base al quale è articolata la correzione della frase, nell'ambito della quale la lookUp è utilizzata ripetutamente e con criteri più maturi al fine di scegliere la soluzione migliore.

### 1.2.2 *LookUp Compound - Intera frase*

Una frase *F* non si corregge semplicemente valutando la *lookUp* delle singole parole che la compongono, su ciascuna di esse è indispensabile condurre una particolare osservazione per garantire il riconoscimento di ortografie in cui uno o più vocaboli sono saldati insieme o al contrario, scritti con separatori che non hanno senso. Il processo si avvia con una fase di tokenizzazione specifica che tiene conto di spazi, punteggiatura e caratteri speciali, il cui solo scopo è la creazione di una bagOfWords di vocaboli, schematizzabile come un vettore *T* di parole: *T = [t$_1$, t$_2$, …, t$_n$]*; la procedura di riconoscimento itera il vettore e si comporta come indicato di seguito.

**Elaborazione del primo elemento.**
$t_1$ è il primo token di F, non ha alcun predecessore e verifica una delle seguenti:

1. E' ortograficamente corretto, la sua lookUp riporta distanza nulla, $t_1$ è un termine di dizionario.
2. Non è presente nel dizionario ma può essere corretto con un vocabolo generatore da cui è distante $d_1$
3. Non è presente nel dizionario ma può essere corretto da una coppia di termini legati da apostrofo o separati da spazio, nel secondo caso la distanza $d_1$ è calcolata sulla coppia e la frequenza sul valore più basso di ciascuno dei due termini.

---

Con $K \in ]0,1]$ parametro stabilito in base alla lunghezza media delle parole presenti nel dizionario, negli esempi di cui faremo menzione da qui in seguito si è scelto di fissare k=0.33; per cui $edit = Int(\frac{1}{3}L)$

**Elaborazioni successive al primo indice.**

$t_i$ è un token di F successivo al primo e verifica una delle seguenti:

1. E' ortograficamente corretto, la sua lookUp riporta distanza nulla, $t_i$ è un termine di dizionario.
2. Non è presente nel dizionario ma può essere corretto con un vocabolo generatore da cui è distante $d_i$, in armonia con il termine che lo precede secondo la verosimiglianza.
3. Non è presente nel dizionario ma se accorpato al termine che lo precede, può essere corretto con un vocabolo generatore da cui è distante $d_i$
4. Non è presente nel dizionario ma può essere corretto da una coppia di termini legati da apostrofo o separati da spazio, nel secondo caso la distanza $d_i$ è calcolata sulla coppia e la frequenza sul valore più basso di ciascuno dei due termini.

La correzione di *F* introduce la necessità di abbinare i token non solo per riconoscerne la corretta ortografia ma anche per stabilire se la sequenza che essi generano è compatibile con le correzioni oppure no e ciò avviene in maniera ricorsiva utilizzando le liste di verosimiglianza. Nel corso del processo generale di valutazione di *F* si evidenziano i due aspetti specifici sui quali intendiamo scendere più in dettaglio:

- *A1: La presenza dell'apostrofo come legante tra articolo/preposizione e sostantivo.* La costruzione del dizionario prevede il censimento delle forme contenenti l'apostrofo come parola singola; ad esempio, nel dizionario dei termini sono presenti *l'acqua, un'acqua, dell'acqua* e *acqua* come voci distinte e quindi elementi generatori della KB. Pertanto l'apostrofo diventa elemento imprescindibile nel riconoscimento di *F* ed entra di diritto nella costruzione delle correzioni.
- *A2: L'individuazione delle strutture significative espresse in nomi composti (ed es: Conto Corrente)* all'interno dei quali confluiscono anche le *diciture rilevanti*, se esistono, del contesto specifico. Non è rara la presenza di pattern costituiti da vocaboli tecnici o gergali, il cui significato è riferibile a concetti di particolare importanza nel dominio di riferimento, se pensiamo ad un operatore telefonico, i fantasiosi nomi delle offerte, che talvolta mescolano italiano e inglese, diventano di fatto, nomi composti la cui esistenza ha senso solo in quell'ambito; ad esempio: "*All inclusive unlimited per lo studio*". L'algoritmo di correzione tiene conto di queste particolarità e le tratta come pattern, nel loro complesso, senza intervenire nella correzione puntuale dei singoli vocaboli che potrebbe alterare il significato complessivo: pensiamo al sintagma "*all inclusive*" su cui si potrebbe tentare la riconoscibilità dell'apostrofo che comunque in questo caso fallirebbe, vista l'impossibilità linguistica della forma *all'inclusive*. L'algoritmo è capace di evidenziare i pattern e isolarli, al fine di segnalarne esplicitamente la presenza, lasciandoli tuttavia al loro posto nella frase corretta.

## 2. Riconoscimento dell'apostrofo (Algoritmo A1)

Intendiamo affrontare il riconoscimento dell'apostrofo in due casi circoscritti che ricoprono la maggior parte delle ortografie errate: parole scritte senza apostrofo in cui è presente uno spazio tra articolo/preposizione e sostantivo (*un amica* -> *un'amica*) oppure termini scritti come un unico vocabolo ma che hanno senso solo se separati da apostrofo (*laltro* -> *l'altro*).

L'algoritmo che proponiamo è basato sul concetto di lookUp (e quindi di distanza minima) di cui si è detto in 1.2.1, cui si aggiungono specifiche manipolazioni delle stringhe al fine di individuare tutte le possibili articolazioni e correzioni dei vocaboli nei quali è rintracciabile la presenza dell'apostrofo. Lo scopo dell'algoritmo non è la correzione ortografica dell'intera frase, ma l'individuazione dei sintagmi che potrebbero coerentemente contenere l'apostrofo. Descriviamo a seguire, i passi di cui si compone il procedimento, indicandone le peculiarità:

1. Tokenizzazione della frase. Il procedimento salva l'apostrofo nel senso che se due termini sono legati tali rimangono: (*tra l'altro* -> [*tra, l'altro*]) cosi come avviene nella costruzione del dizionario. Naturalmente si escludono dal flusso quei token che già contengono l'apostrofo.
2. Iteriamo sui token ispezionabili.
3. *Primo termine*: il procedimento consiste nell'introdurre l'apostrofo nel token tra il primo e il secondo carattere e farlo *navigare* verso destra fino ad arrivare al penultimo carattere, si ottengono cosi determinazioni come quelle in tabella:

| dllaltro | |
|---|---|
| 1 | d'llaltro |
| 2 | d'llaltro |
| 3 | dl'laltro |
| 4 | dll'altro |
| 5 | dlla'ltro |
| 6 | dllal'tro |
| 7 | dllalt'ro |
| 8 | dllaltr'o |

*Tabella 2.1 - Drift dell'apostrofo*

Si esegue la lookUp su ciascuna determinazione e se esistono correzioni valide si collezionano, la correzione che possiede distanza minima sostituirà il token in *F*. Nella tabella sopra, la determinazione 4 è quella che genera la correzione *dell'altro* con distanza minima (pari a 1). Se non esistono correzioni valide, il termine corrente non è idoneo a contenere l'apostrofo e si passa alla successiva iterazione.

4. *Termini successivi al primo*: la prima azione consiste nel verificare che l'elemento corrente $t_n$ e il precedente $t_{(n-1)}$ costituiscano una coppia linguisticamente valida; e a tale scopo si utilizza la lista di verosimiglianza delle coppie estratte dal corpus. L'ispezione procede ciclicamente accoppiando le correzioni di $t_{(n-1)}$ con ciascuna correzione di $t_n$. Siano $[t_{(n-1)}]$ e $[t_n]$ le classi di correzione generate dai rispettivi elementi: e diciamo $c_{(n-1)}$ e $c_n$ due rappresentati generici di quelle classi, se esiste una coppia ammissibile, linguisticamente valida, che armonizzata mediante introduzione dell'apostrofo cioè $c_{(n-1)}'c_n$, ottiene una lookUp a distanza nulla, allora il termine $c_{(n-1)}'c_n$ sostituisce in *F* la coppia ($t_{(n-1)}$, $t_n$). E' possibile ridurre il numero di coppie su cui eseguire i controlli: non si considerano i vocaboli che pur essendo ammissibili in sequenza, non hanno senso in un sintagma contenente l'apostrofo; è il caso in cui $c_{(n-1)}$ termina con una vocale e $c_n$ inizia con una consonante, ad esempio (*la, strada*) non prevede apostrofo mentre (*dell, acqua*) lo prevede. È particolarmente importante osservare che: **se i due elementi di partenza sono corretti** (a distanza nulla quindi vocaboli di dizionario) il corrispondente sintagma contenente l'apostrofo sostituisce la coppia se e solo se è un termine del dizionario. Ad esempio la coppia (*l, acqua*) produce il sintagma *l'acqua*, linguisticamente ammissibile e presente nel dizionario; la coppia (*le, acque*) genera la forma inesistente *le'acque* che non è termine di dizionario ma suscettibile di correzione ortografica (con *l'acqua*), quindi non sostituisce la coppia di partenza che rimane correttamente invariata in *F*. Nel caso in cui non si pervenga ad alcun tipo di arricchimento, si procede con l'indagine come al punto 3 al fine di verificare se $t_n$ può contenere o meno un apostrofo.

# 3. Riconoscimento delle espressioni polirematiche (Algoritmo A2)

Il riconoscimento delle espressioni polirematiche (*EP*) è possibile creando un'adeguata struttura dati che coniuga il dizionario generale di sistema *D* (mutuato da Paisà) con uno specifico elenco di espressioni composte $E_c$ che comprenda quelle generali di uso comune (*Conto Corrente, Cassa di risonanza*, ecc...) e quelle specifiche appartenenti al contesto di riferimento. L'algoritmo lavora sull'unione dei due dizionari:

$$G = D \cup E_c$$

Sia *F* una frase che contiene una o più espressioni polirematiche:

    *F* = *aspettiamo la risposta della cmnità erupea e della presidenza del cosniglio dei ministri*    (1)

Per garantire una migliore comprensione del testo, definiamo alcune misure:

- $L_F$ *Lunghezza della frase*:     E'il numero di token che compongono *F*
- $L_c$ *Lunghezza della Composta*:     E'il numero di token che compongono la singola espressione composta

Da un punto di vista formale l'intera *F* potrebbe essere una EP quindi $L_F = L_c$ o contenere al suo interno da *0* a *n* EP. Riconosciamo che la (1) contiene evidentemente due EP. A priori non è possibile stabilire quali sequenze di termini in *F* individuano una EP, per cui l'algoritmo procede costruendo liste di combinazioni di termini a partire dalla collezione dei token estraibili da *F*; la lunghezza minima ammissibile per una combinazione è 1 token[3], quella massima coincide con $L_F$ o con il max($L_c$) in G. In generale possiamo dire che detta n la dimensione della n-pla si ha: $n \in [1, Sup]$ in cui:

$$Sup = \min (L_F, \max(L_C)) \quad (3.1)$$

Riepilogando:

1) Tokenizzazione di *F* cosi come avviene nella costruzione del dizionario.
2) Costruzione di liste di n-ple di token, con *n* in [1, Sup].

L'esistenza delle n-ple e quindi delle corrispondenti lookUp mirate a individuare le espressioni polirematiche coinvolte in *F*, consente la creazione del catalogo di EP associato a *F*, l'algoritmo utilizza ciascuna EP come chiave di accesso alla n-pla che in *F* rappresenta l'espressione che sia ortograficamente corretta oppure no.

Spieghiamo il procedimento utilizzando l'esempio offerto dalla (1). Saltiamo il dettaglio per n=1 perché in quel caso specifico nessun elemento singolo individua una EP, per n = 2 si ottiene:

| n = 2 | |
|---|---|
| 1 | aspettiamo la |
| 2 | la risposta |
| 3 | risposta dlla |

---

[3] Il token singolo può rappresentare una polirematica in quanto l'ipotesi sotto la quale svolgiamo questo lavoro è la probabilità di ortografia non corretta, nella quale i singoli termini potrebbero essere fusi insieme: *cmnitàerupea*

| 4  | dlla cmnità       |
|----|-------------------|
| 5  | *cmnità erupea*   |
| 6  | erupea e          |
| 7  | e della           |
| 8  | della presidenza  |
| 9  | presidenza del    |
| 10 | del cosniglio     |
| 11 | cosniglio dei     |
| 12 | dei ministri      |

*Tabella 3.1 - Elenco delle combinazioni possibili per n = 2*

Ogni n-pla è candidata ad essere una EP, per escludere o confermare l'ipotesi si esegue una lookUp su ciascuna di esse; quelle che effettivamente rappresentano un EP si raccolgono in una collezione e in F si procede a sostituire la n-pla con la correzione che possiede la distanza minore; dalla coppia 5 otteniamo:

| comunità europea | | |
|---|---|---|
| n = 2 | Token | *cmnità erupea* |
|       | Distanza | 4 |

*Tabella 3.2 - Risultato finale per **comunità europea***

da nessuna delle altre n-ple si ottiene la stessa EP con distanza minore, all'interno di F la sequenza: *cmnità erupea* può essere sostituita dalla corrispondente espressione polirematica: *comunità europea*.

Per quanto concerne la seconda espressione la faccenda si complica, perché vi sono collezioni di n-ple che individuano due diverse EP, stante la circostanza che in G è presente tanto "*Presidenza del Consiglio*", quanto "*Presidenza del Consiglio dei Ministri*".

In dettaglio si hanno le 3-ple (e analogamente le 4-ple e le 5-ple):

| n = 3 | |
|----|---------------------------|
| 1  | aspettiamo la risposta    |
| 2  | la risposta dlla          |
| 3  | risposta dlla cmnità      |
| 4  | dlla cmnità erupea        |
| 5  | cmnità erupea e           |
| 6  | erupea e della            |
| 7  | e della presidenza        |
| 8  | della presidenza del      |
| 9  | *presidenza del cosniglio* |
| 10 | del cosniglio dei         |
| 11 | cosniglio dei ministri    |

*Tabella 3.3 - Elenco delle combinazioni possibili per n = 3*

Da cui l'algoritmo genera la sequenza:

| presidenza del consiglio | | |
|---|---|---|
| n = 3 | Token | *presidenza del cosniglio* |
| | Distanza | 1 |
| n = 4 | Token | *presidenza del cosniglio dei* |
| | Distanza | 5 |
| n = 4 | Token | *della presidenza del cosniglio* |
| | Distanza | 7 |
| n = 5 | Token | *e della presidenza del cosniglio* |
| | Distanza | 9 |
| n = 5 | Token | *della presidenza del cosniglio dei* |
| | Distanza | 11 |

*Tabella 3.4 - Risultato finale per **presidenza del consiglio***

L'espressione *presidenza del consiglio* è *raggiungibile* da più n-ple (3, 4, 5) ma solo quella che corrisponde a n = 3 ha distanza minima, per cui il corrispondente token potrebbe essere sostituito in F; ma l'elaborazione non è ancora terminata infatti per n = 5 si ha:

| n = 5 | |
|---|---|
| 1 | aspettiamo la risposta dlla cmnità |
| 2 | la risposta dlla cmnità erupea |
| 3 | risposta dlla cmnità erupea e |
| 4 | dlla cmnità erupea e della |
| 5 | cmnità erupea e della presidenza |
| 6 | erupea e della presidenza del |
| 7 | e della presidenza del cosniglio |
| 8 | della presidenza del cosniglio dei |
| 9 | presidenza del cosniglio dei ministri |

*Tabella 3.5 - Elenco delle combinazioni possibili per n = 5*

| presidenza del consiglio dei ministri | | |
|---|---|---|
| n = 5 | Token | *presidenza del cosniglio dei ministri* |
| | Distanza | 1 |
| n = 5 | Token | *della presidenza del cosniglio dei ministri* |
| | Distanza | 7 |

*Tabella 3.6 - Risultato finale per **presidenza del consiglio dei ministri***

Da ciò si evince che il token: presidenza del cosniglio dei ministri può essere sostituito da presidenza del consiglio dei ministri. L'algoritmo sostituisce in F i token che corrispondono alle EP collezionate nella fase di costruzione delle n-ple. Il caso di inclusione come quello evidenziato nell'espressione [[*Presidenza del Consiglio*] *dei ministri*] si risolve applicando un tipo di sostituzione incrementale tanto in F quanto nei token associati alle EP individuate durante l'elaborazione:

| Sostituzioni in *F* | | | |
|---|---|---|---|
| Passo 1 | Token | | *presidenza del cosniglio* |
| | $EP_1$ | | **Presidenza del consiglio** |
| | F | | *aspettiamo la risposta della cmnità erupea e della $EP_1$ dei ministri* |
| Passo 2 | L'espressione $EP_1$ si sostituisce al token in tutte le sue occorrenze, quindi anche in Tabella 3.6 che assume la forma: | | |
| | **presidenza del consiglio dei ministri** | | |
| | n = 5 | Token | *$EP_1$ dei ministri* |
| | | Distanza | 1 |
| | n = 5 | Token | *della $EP_1$ dei ministri* |
| | | Distanza | 7 |
| | In F si sostituisce il token valido nella tabella sopra. | | |
| Passo 3 | Token | | *$EP_1$ dei ministri* |
| | $EP_2$ | | **Presidenza del consiglio dei ministri** |
| | F | | *aspettiamo la risposta della cmnità erupea e della $EP_2$* |
| Passo 4 | Token | | *cmnità erupea* |
| | $EP_3$ | | **Comunità europea** |
| | F | | *aspettiamo la risposta della $EP_3$ e della $EP_2$* |

*Tabella 3.7 - Sostituzione delle espressioni polirematiche*

Il risultato finale dell'elaborazione propone la frase F corretta, con l'evidenza completa delle espressioni polirematiche in essa contenute:

| F | *aspettiamo la risposta della comunità europea e della presidenza del consiglio dei ministri* |
|---|---|
| $EP_2$ | ***presidenza del consiglio dei ministri*** |
| $EP_3$ | ***comunità europea*** |

## 4. Bibliografia